%% file: main.tex
\documentclass[conference]{IEEEtran}
\IEEEoverridecommandlockouts
\usepackage{cite}
\usepackage{amsmath,amssymb,amsfonts}
\usepackage{algorithmic}
\usepackage{graphicx}
\usepackage{textcomp}
\usepackage{xcolor}
\usepackage{soul}
\usepackage[a4paper, total={184mm,239mm}]{geometry}
\def\BibTeX{{\rm B\kern-.05em{\sc i\kern-.025em b}\kern-.08em
    T\kern-.1667em\lower.7ex\hbox{E}\kern-.125emX}}

\renewcommand{\baselinestretch}{0.975}

\usepackage{tikz}
\usepackage{textcomp}
\usepackage[doipre={DOI:~}]{uri}
\usepackage{lipsum}
\newcommand\copyrighttext{%
    \footnotesize \textcopyright 2024 IEEE. Personal use of this material is permitted.  Permission from IEEE must be obtained for all other uses, in any current or future media, including reprinting/republishing this material for advertising or promotional purposes, creating new collective works, for resale or redistribution to servers or lists, or reuse of any copyrighted component of this work in other works.
    
    Accepted as a conference paper at the 2025 Design, Automation and Test in Europe Conference.
    }
\newcommand{\copyrightnotice}{%
\begin{tikzpicture}[remember picture,overlay]
\node[anchor=south,yshift=10pt] at (current page.south) {\fbox{\parbox{\dimexpr\textwidth-\fboxsep-\fboxrule\relax}{\copyrighttext}}};
\end{tikzpicture}%
}

\begin{document}
\bstctlcite{IEEEexample:BSTcontrol}
\title{Coupling Neural Networks and Physics Equations For Li-Ion Battery State-of-Charge Prediction
\thanks{This publication is part of the project PNRR-NGEU which has received funding from the MUR – DM 117/2023.}
}
\input{text/authors}
\maketitle
\copyrightnotice
\input{text/00-abstract}

\vspace{-0.5cm}
\input{text/01-introduction}
\input{text/02-background}
\input{text/03-methods}

\input{text/04-results}

\input{text/05-conclusions}
\newpage
\bibliographystyle{IEEEtran}
\bibliography{references}

\end{document}

%% file: text/authors.tex
\author{\IEEEauthorblockN{
Giovanni Pollo, Alessio Burrello, Enrico Macii, Massimo Poncino, Sara Vinco, Daniele Jahier Pagliari}
\IEEEauthorblockA{ 
Politecnico di Torino, Turin, 10129, Italy}
\IEEEauthorblockA{Emails: name.surname@polito.it}
}

%% file: text/00-abstract.tex
\begin{abstract}
Estimating the evolution of the battery's State of Charge (SoC) in response to its usage is critical for implementing effective power management policies and for ultimately improving the system's lifetime.
Most existing estimation methods are either physics-based digital twins of the battery or data-driven models such as Neural Networks (NNs).
In this work, we propose two new contributions in this domain. First, we introduce a novel NN architecture formed by two cascaded branches: one to predict the \textit{current} SoC based on sensor readings, and one to estimate the SoC at a \textit{future} time as a function of the load behavior. Second, we integrate battery dynamics equations into the training of our NN, merging the physics-based and data-driven approaches, to improve the models' generalization over variable prediction horizons. 
We validate our approach on two publicly accessible datasets, showing that our Physics-Informed Neural Networks (PINNs) outperform purely data-driven ones while also obtaining superior prediction accuracy with a smaller architecture with respect to the state-of-the-art.

\end{abstract}

%% file: text/01-introduction.tex
\section{Introduction}
Safe and efficient operations of battery-powered devices at any scale (from portable electronics to electric vehicles) require advanced Battery Management Systems (BMS) in order to monitor the internal state of a battery and optimize its usage.
Among the crucial parameters monitored by the BMS, the SoC, %
which measures the remaining battery charge, plays a critical role in ensuring battery longevity and averting potential failures.
Moreover, the SoC serves as an input for the calculations of other quantities, such as the State-of-Health (SoH), battery power, and cell balancing.

The accurate and reliable estimation of the current SoC and the prediction of its future value in response to specific stimulation of the battery is a challenging task due to its dependence on factors like battery age, manufacturing variability, ambient temperature, etc. As a matter of fact, previous works have even claimed that it is virtually impossible to measure all the actual electrochemical factors that affect the SoC \cite{REZA14}.
Therefore, purely \textit{physics-based} models that try to relate the SoC to quantities that affect it (or may affect it) are not easy to devise~\cite{8843918}. 
For this reason, a number of \textit{data-driven} approaches have recently appeared, which are essentially Machine Learning (ML) models %
trained offline on easily measurable quantities (i.e., current, voltage, and temperature) to approximate the mathematical relation between them and/or some statistical aggregation thereof, and a ground truth SoC value~\cite{Vidal20}.

Data-driven models have some key advantages over physical ones.
Firstly, not being linked to the physics of a specific type of battery, they are more general and flexible and can adapt to various chemistries, battery sizes, etc., as long as a corresponding training dataset is available.
Secondly, they are usually more efficient in terms of latency, memory footprint, and energy consumption
\cite{10244587}. In particular, they do not require simulation to estimate or predict the SoC, and are therefore particularly suited for resource-constrained BMSs.

This work introduces a novel data-driven method, \textit{especially tailored for the prediction of future SoC values} in response to the application of a specific load profile to the battery, a task that is important for implementing smart or predictive power management policies. Our approach is based on a novel %
Neural Network (NN) architecture that uses two cascaded branches to estimate current SoC and to predict its future behavior, respectively. Furthermore, we make the training of this NN \emph{physics-informed} by adding a loss function term that encourages generalization across prediction horizons. 
In detail, our main contributions are:
\begin{itemize}
    \item We propose an innovative two-branch NN structure where one branch accurately estimates the SoC at time $t$ based on the measured voltage, current, and temperature, and the other predicts the SoC at time $t+N$ (with $N$ variable) given estimated current and temperature profiles.
    \item We embed a physics equation that models \textit{Coulomb counting} in the loss function of the second NN branch (i.e., the predictive one) to regularize the SoC prediction based on the integral current flowing from/to the battery. 
    \item With experiments on two publicly available datasets, Sandia~\cite{Preger_2020} and LG~\cite{Kollmeyer2020}, we demonstrate that the physics-informed loss term enforces better generalization across prediction horizons (i.e., values of $N$) in our model. For instance, when trained with $N=30s$ and tested with $N=70s$, the physics-informed NN reduces the Mean Absolute Error (MAE) by 82\% with respect to a purely-data driven method.
    \item Additionally, on the only dataset for which comparison baselines are available, we demonstrate that our novel two-branch NN achieves comparable results with respect to the LSTM proposed in~\cite{Dang23} (0.014 vs 0.012 MAE), while also allowing SoC prediction, and requiring 409$\times$ fewer parameters.
\end{itemize}

%% file: text/02-background.tex
\vspace{-0.2cm}
\section{Background and Related Works}
\label{sec:background}
The SoC of an energy storage device is defined as the ratio of the available capacity $Q(t)$ and the maximum possible charge that can be stored into it, i.e., $\mbox{SoC}(t) = Q(t)/Q_{max}$.
In spite of its straightforward definition, 
accurate estimation of SoC is non-trivial.
$Q_{max}$ is typically %
assumed to be the nominal battery capacity as provided by the manufacturers, which might not be an accurate guess due to various variability effects \cite{DAHMARDEH2021103204}. Moreover, $Q_{max}$ 
is not constant throughout the battery life due to aging \cite{10244587}.
Because of these and other subtle factors, accurate SoC estimation remains a challenging problem to solve, thereby inspiring a wealth of literature on the topic.
The landscape of SoC estimation methods is so vast that there exists a number of surveys that provide excellent overviews of the various solutions~\cite{8843918,Vidal20}. 

As a compact summary, SoC estimation methods can be broadly classified in three categories:
\begin{enumerate}
    \item  \textit{Direct measurements}, which use some measurable quantity (voltage, resistance/impedance, current) that can be correlated to SoC. This includes methods based on open circuit voltage (OCV) \cite{Ng08}, impedance \cite{Coleman07}, and \textit{Coulomb counting}, in which the SoC is estimated by integrating the discharging current over time \cite{Ng2009}; 
    \item  \textit{Physics-based} approaches, which model the battery by following or approximating the underlying physics; these include electro-chemical models~\cite{westerhoff}, electrical-circuit equivalent models~\cite{islped13}, and models based on state estimation (e.g., Kalman filters)~\cite{kalman};
    \item \textit{Data-driven} approaches, which also rely on a model that is, however, directly extracted from a dataset of current, voltage, and temperature measurements associated to ground-truth SoC values~\cite{lstm_tie,fnn_chemali,Vidal20}; unlike the previous class, such models are not based on physics, and their parameters are rather obtained through a fitting (or training) procedure.
\end{enumerate}
When addressing the more complex problem of SoC \textit{prediction} (i.e., not just estimating the \textit{current} SoC but rather \textit{predicting its future value} in response to how the battery will be stimulated), direct measurements as in (1) are unfeasible, and the only available options are either to build a physics-based digital twin of the battery using a model from category (2), or to train a data-driven forecasting model as in (3).
For the former option, we refer the reader to the above-mentioned surveys for a detailed analysis. %
In the rest of the section, we overview the most relevant data-driven models, which are the main focus of our work.
\vspace{-0.1cm}
\subsection{Data-Driven SoC Models}
\vspace{-0.1cm}
A number of different data-driven solutions have been proposed to estimate battery states (SoC and/or SoH). They differ essentially in the structure and complexity of the ML model adopted, ranging from simple classic ones such as tree-based ensembles~\cite{10244587} to deep feedforward~\cite{fnn_chemali} or recurrent NNs~\cite{wong-2021}. The survey of \cite{Vidal20} provides an exhaustive overview of ML-based approaches for battery state estimation.

In this domain, the problem of the generalization of ML models (i.e., their ability to accurately model data different from those seen during training) is particularly critical. In fact, in-field data might correspond to widely varying system workloads (i.e., current and temperature profiles), while most training datasets are obtained by controlled lab measurements. However, recent studies have observed that by incorporating prior information (e.g., physical laws or domain knowledge) in the learning process, it is possible to enhance the generalization of the models and also accelerate training~\cite{Raissi21}.

Physics-informed neural networks (PINNs) are an example of this learning bias, in which the physics of the underlying phenomenon is added to the model's training loss function as penalty constraint in the form of partial differential equations (PDEs) \cite{10.1007/s10915-022-01939-z}.
A few recent works have used physics-informed ML models for SoC estimation \cite{Tian22,Wang22,Ahn23,Dang23},
by incorporating some form of battery dynamics inside the model.
Our work is closest to \cite{Dang23} in that it is the only ``true'', canonical PINN architecture of the above list; the authors include the PDEs corresponding to the dynamics of a first-order circuit-equivalent RC battery model into the loss function of conventional NN models, namely a Multi-Layer Perceptron (MLP), and two types of recurrent NNs.
However, our work differs from \cite{Dang23} in several aspects. First, similarly to \cite{Bian2020}, \textit{we target SoC prediction rather than estimation}, although our proposed model is capable of both. As a second element of novelty, we propose a novel NN architecture formed by two cascaded branches: one to predict the current SoC based on sensor readings and one to estimate the SoC at a future time as a function of the expected battery usage. This architecture is significantly smaller and less computationally expensive than the ones in~\cite{Dang23}.
Third, we incorporate physics equations in our model's loss function with a different purpose, that is, to enhance its generalization in predicting the future SoC at \textit{multiple time horizons}.

%% file: text/03-methods.tex
\section{Methodology}
\label{sec:methodology}
Accurately predicting the future SoC in response to a current profile drawn from the battery enables several advanced power management optimizations. On an electric vehicle, either terrestrial or aerial (e.g., a small drone), it allows taking runtime decisions on the best route to follow to maximize battery lifetime~\cite{tvt}. On a battery-operated embedded device, it could be used to find the most appropriate scheduling of computing tasks~\cite{glsvlsi}. Being able to perform this prediction with \textit{multiple time horizons} enables the combination of faster-yet-approximate long-term decisions (e.g., on the best overall route) with slower-yet-precise short-term ones (e.g., on the optimal speed and altitude for the next route segment).
In the rest of this section, we present a data-driven SoC prediction method that favors such multi-horizon applications. Namely, we first introduce our proposed NN structure and then describe how we enhance its training using a physics law. 
\subsection{Network Architecture}\label{sec:arch}
\begin{figure}[t]
    \centering
    \includegraphics[width=.99\linewidth]{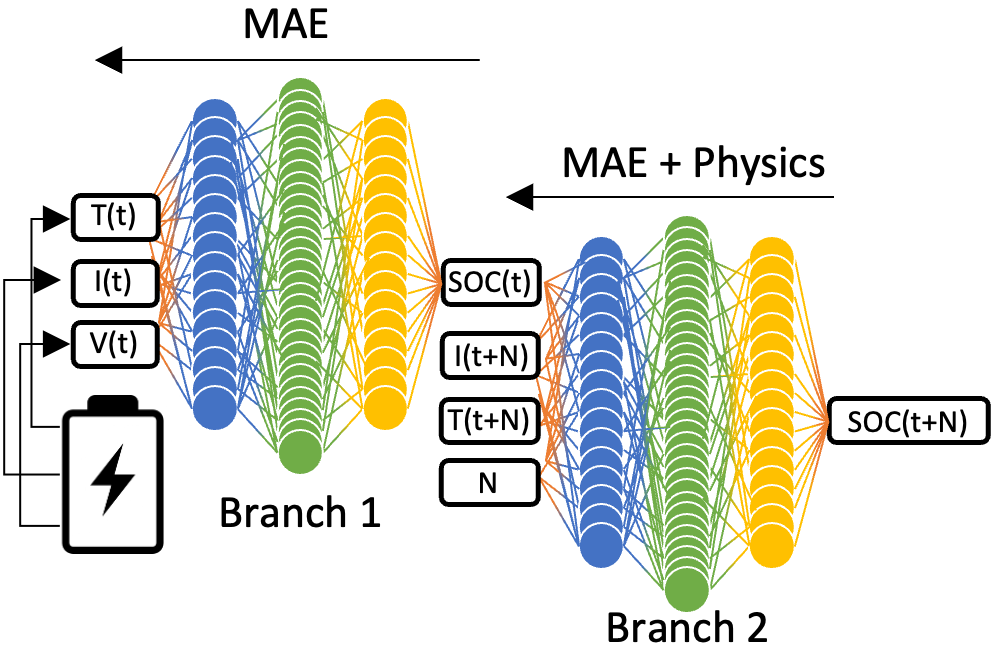}
    \vspace{-0.7cm}
    \caption{Proposed two-stage neural network architecture.}
    \vspace{-0.5cm}
    \label{fig:nn-structure}
\end{figure}

\vspace{-0.1cm}
Figure \ref{fig:nn-structure} depicts the architecture of our proposed NN. The overall model is formed by cascading together two Fully-Connected (FC) feed-forward sub-networks, or branches. 

\emph{Branch 1} takes as inputs easily measurable (and available in any public dataset) information on the state of the battery at time $t$, namely: the voltage $V(t)$, the drawn current $I(t)$ and the temperature $T(t)$. Its output is an estimation of the SoC at the same time instant, i.e., $SoC(t)$. Notice that all three inputs are required since both current and temperature influence the instantaneous SoC, e.g., through the voltage drops across internal battery resistances.

\emph{Branch 2} has the objective of predicting the future value of the SoC in response to how the battery is used; its output is $SoC(t+N)$, where  $N$ denotes the \textit{time horizon} of the prediction.
Branch 2 takes the output of Branch 1 as input and uses it as \textit{initial condition} for the prediction. 
In addition to the estimated initial SoC, Branch 2 receives three additional inputs, representing the workload for which we want to estimate the future SoC.
These are the \textit{average} current $I(t+N)$ applied between $t$ and $t+N$, the corresponding average temperature $T(t+N)$, and the prediction horizon $N$.

When querying the second model, these three inputs serve as user-specified parameters;
while the inputs of Branch 1 are measurable quantities, $I(t+N), T(t+N)$, and $N$ are used to define \textit{the workload conditions and the temporal horizon for which we want to calculate the future SoC, based on the current $SoC(t)$}.
While $I(t+N)$ represents a hypothetical workload, estimating the future temperature $T(t+N)$ is not trivial. However, for a relatively short horizon, it is reasonable to assume constant temperature without significant losses of accuracy, i.e., setting $T(t+N) = T(t)$.

The prediction horizon $N$ is needed as input because the amount of time elapsed is obviously strictly correlated with the SoC variation. Thus, adding it as an external input, rather than letting the model infer the elapsed time from data, allows us to have a single NN that can generate predictions at multiple instants in the future, which is fundamental to support multi-horizon power management as discussed above.

\begin{figure}[ht]
    \centering
    \includegraphics[width=.99\linewidth]{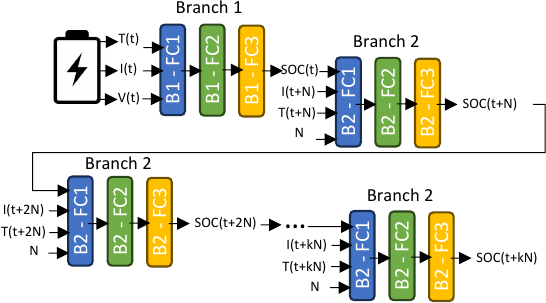}
    \vspace{-0.5cm}
    \caption{Multi-step autoregressive prediction with the proposed NN architecture.}
    \vspace{-0.2cm}
    \label{fig:nn-autoregressive}
\end{figure}

Figure~\ref{fig:nn-autoregressive} shows how the proposed model can be used to estimate the SoC over multiple future timesteps with non-constant current requests. Here, the NN layers have been schematized as boxes for simplicity, where Bx-FCy indicates the $y$-th FC layer of the $x$-th branch. As shown, after executing Branch 1 a single time to estimate the initial $SoC(t)$, $k$ consecutive executions of Branch 2 in an autoregressive fashion (i.e., feeding the output of the $i$-th invocation as input for the $i+1$-th) are used to obtain a complete estimate of the SoC evolution in response to the battery's stimulation pattern. %

In this work, we use the same hyper-parameters for the two branches (except for the number of inputs, set to 3 and 4, respectively). Namely, each branch has 3 FC hidden layers with ReLU activation functions and 16, 32, and 16 units, respectively, following an inverted bottleneck structure. The output layer has a single unit in both cases, with no activation, to predict an unrestricted scalar value. While we leave the exploration of the NN architecture to future work, we remark that this model requires a pretty small number of trainable parameters (2,322 corresponding to approximately 9kB of storage with {\tt float32} representation), thus being suited for performing low-cost runtime predictions on-board a BMS or a Power Management Integrated Circuit (PMIC).

\subsection{Training Scheme and Loss Function}\label{sec:loss}
The network architecture described in Sec.~\ref{sec:arch} can be trained in a purely data-driven fashion. In that case, both branches use the Mean Absolute Error (MAE) between the predicted and ground truth SoC as a loss function to be minimized via gradient descent. We empirically observed that training the two branches separately, i.e., stopping the back-propagation of gradients from Branch 2 to Branch 1, yields superior results. In other words, during training, the weight updates of Branch 1 only depend on the error on the prediction of $SoC(t)$, and not on the propagated error from Branch 2. Moreover, exclusively at training time, Branch 2 is fed with ground truth $SoC(t)$ values from the dataset (whereas at inference times it receives the estimated SoC from Branch 1). This split training also improves the explainability of our model, as Branch 1 outputs a physically relevant quantity ($SoC(t)$) rather than a black-box intermediate value. 

To enhance the purely data-driven setup and improve the network's generalization, Branch 2 can be turned into a PINN, adding a physics component to its loss function. In particular, in our work, we use the simple \textit{Coulomb counting} equation, which associates the SoC difference over time to the total net charge exchanged (provided or absorbed) by the battery. Mathematically, the equation is the following:
\vspace{-0.15cm}
\begin{equation}
    \label{eq:cc}
    SoC_p(t+N_p) = \frac{1}{C_{rated}} \int_{t}^{t+N_p} I(t)dt + SoC(t)
\end{equation}
where subscript $p$ stands for ``physics", and $C_{rated}$ is the battery's nominal capacity, as reported in the datasheet. 
When using this setup, the overall loss function of Branch 2 becomes:
\vspace{-0.6cm}
\begin{multline}
    \mathcal{L} = MAE(SoC(t+N), \hat{SoC}(t+N)) + \\
    MAE(SoC_p(t+N_p), \hat{SoC}(t+N_p))
\end{multline}
where $\hat{SoC}$ indicates the NN prediction. 

During training, in correspondence with each minibatch of data used to evaluate the first loss component, the physics-based loss is computed over a set of \textit{different, randomly generated} values of initial SoC, current, and time delta conditions. 
In particular, the time delta for this second term ($N_p$) takes multiple values drawn from a set $\mathcal{N}$, which are in general different from the fixed value of $N$ used for the data-driven part, which is constrained by the sampling frequency of the dataset. Conversely, $\mathcal{N}$ includes smaller and/or larger values, thus enabling the model to learn how to predict the SoC degradation at multiple future instants simultaneously.  

While Eq.~\ref{eq:cc} clearly neglects many secondary effects (thermal, cell-to-cell variability, etc.), it still enforces the first-order behavior of the SoC evolution as a function of the requested current, thus acting as a \textit{regularization component} in the loss. 
This has the effect of improving the NN performance on unseen data, even when using a value of $N$ different from the ones applied during training.

It has to be noted that, currently, our model does not account for battery SoH degradation. Therefore, it is accurate only for relatively short horizons (e.g., hours, not months), and only as long as the actual SoH is comparable to the one of batteries included in the training set. While out-of-scope for this paper, one simple way to cope with this limitation, and make sure that the proposed system remains accurate across varying SoH conditions, follows the approach of~\cite{coins}. There, the authors build an ensemble of SoC prediction models, each trained with data at a different SoH level, and select the appropriate one to use based on a separate SoH estimation model.
\section{Datasets}\label{sec:datasets} 
\vspace{-0.1cm}
\subsection{Sandia}
\vspace{-0.1cm}
This dataset, collected by the Sandia National Lab \cite{Preger_2020}, contains charge and discharge cycles of three different 18650
commercial batteries (NCA, NMC, and LFP). 
The batteries are charged and discharged using different currents until the end of their capacity, leading to cycles with different durations.
The charge/discharge current ranges from 0.5C to 3C, and the temperature from 15°C to 35°C. The data are sampled every 120 seconds.
In our benchmarking, we consider all the data with a charge/discharge current of 0.5C/-1C to train the network and with a charge/discharge current of 0.5C/-2C and 0.5C/-3C to test the network.

We consider SoC predictions over a time horizon of N = 120s, i.e., the time delta between two consecutive samples in the dataset, for the data-driven loss component of Branch 2.
For the physics loss, we generate an identical number of cycles, with the same current conditions of the dataset, and prediction horizions $N_p$ set to 120s, 240s, 360s, or to an even mix of all three. 
We identify the corresponding trained PINNs as PINN-120s, PINN-240s, PINN-360s, and PINN-All, respectively.

It's important to note that
for the physics loss,
labels are not necessary, as future SoC values come directly from Eq. \ref{eq:cc}. This is a significant advantage of the PINN, as it allows the network to be trained across \textit{any} time horizon
(longer or shorter than the one of the data),
without relying on ground truth labels, unlike traditional data-driven methods. In our experiments, we
limit ourselves to values of $N_p \ge N$ just because it allows us to later \textit{test the models in those conditions} by under-sampling the dataset, as detailed below (whereas testing with $N_p < N$ would be impossible).

We assess the goodness of the trained Branch 1 (SoC estimation part) by predicting SoC(t) on the unseen dataset cycles and computing the MAE w.r.t. the ground truth. For testing the whole model (Branch 1 + Branch 2), in order to demonstrate the advantage of our PINN, we consider three test sets, targeting the prediction of: SoC(t+120s), SoC(t+240s), and SoC(t+360s), where the former is directly obtained from consecutive dataset samples, and the other two are obtained by taking sliding windows of the dataset, averaging current and temperature values in each window, and using the final SoC value as prediction target.
\subsection{LG}
This dataset was collected at McMaster University \cite{Kollmeyer2020} on an LGHG2 3Ah battery. Contrary to Sandia, charge/discharge cycles collected in this dataset are not characterized by a constant current but are created using specific patterns belonging to different driving conditions. Specifically, the patterns considered are the UDDS (Urban Dynamometer Driving Schedule), the HWFET (Highway Fuel Economy Test), the LA92, and the US06. 
Temperatures are in the -20°C to +40°C range. The sampling rate is set to 0.1s, and a single cycle is present for each driving condition. Furthermore, eight charge/discharge cycles composed of a mixture of the four patterns are also collected.
As done in \cite{wong-2021}, we selected seven out of eight mixed cycles as training data, with temperatures ranging from 0°C to 25°C. For testing, we used the remaining four cycles, each representing a different driving pattern, along with the final mixed cycle. %

In this case, due to the higher granularity of the samples, we decided to use shorter time horizons of 30s, 50s, and 70s for testing.
Before feeding the data to our neural network, we added a moving average of 30s as pre-processing that smooths the I, V, and T values and removes noisy peaks that could arise from the fine-grained sampling of this dataset.
Both the physics data and the test conditions have been generated identically to the Sandia dataset, except for the different time horizons.

%% file: text/04-results.tex
\vspace{-0.1cm}
\section{Experimental Results}
\vspace{-0.1cm}
\label{sec:results}

\begin{figure}[t]
    \centering
    \includegraphics[width=0.99\linewidth]{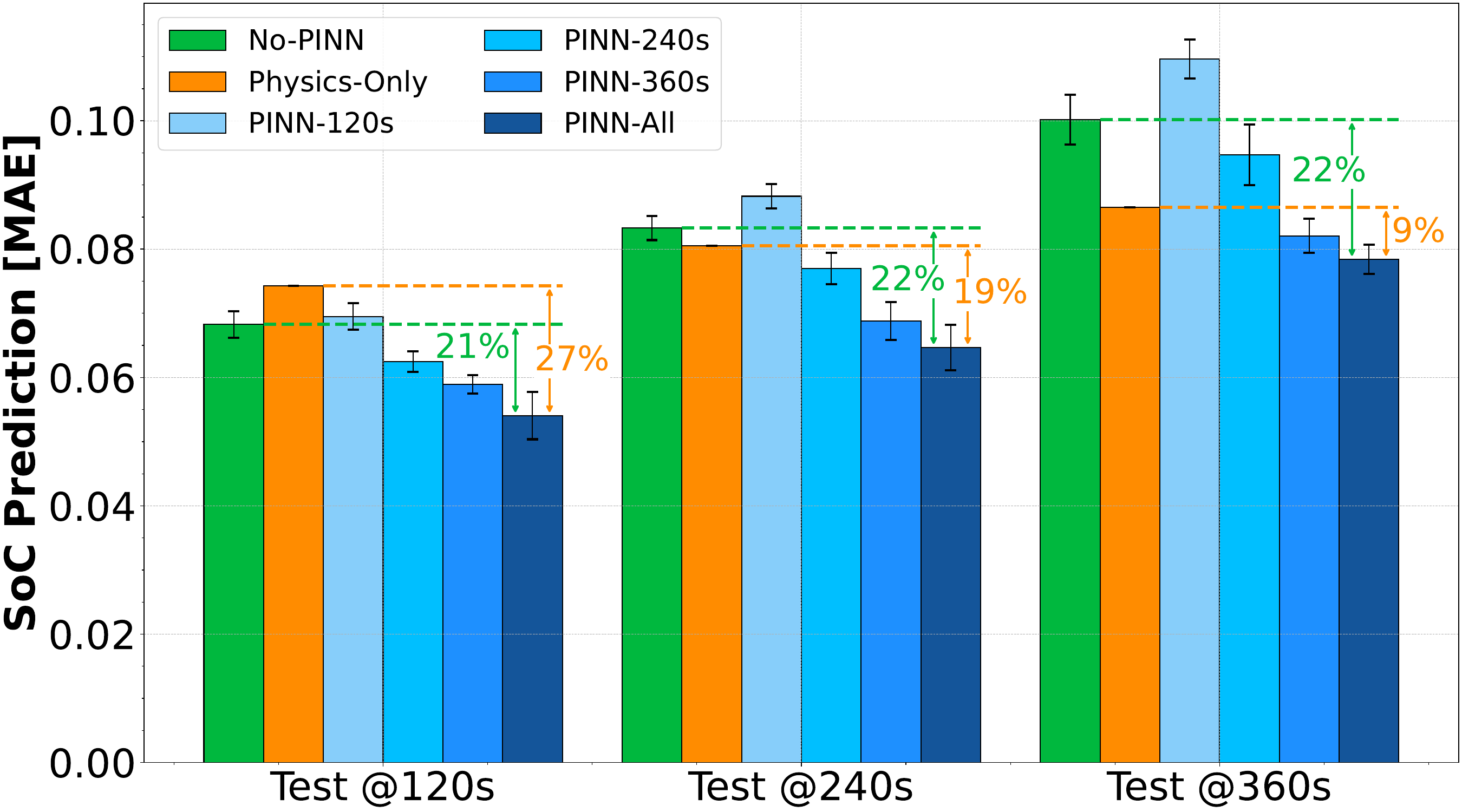}
    \vspace{-0.5cm}
    \caption{Results on Sandia dataset with different physic loss}
    \vspace{-0.5cm}
    \label{fig:sandia}
\end{figure}
\subsection{Results on the Sandia Dataset}
\vspace{-0.1cm}
We first benchmark our proposed network on the Sandia Dataset, showing the network performance on the three test conditions described in the previous section.
Figure \ref{fig:sandia} shows the overall results in terms of SoC prediction MAE. 
The x-axis groups the results by test dataset, each corresponding to a different time horizon (120s, 240s, and 360s, respectively). Within each group, each of the 6 bars represents a specific configuration of the physics loss. Physics-Only is a model whose second branch predicts the SoC \textit{exclusively} employing Eq.~\ref{eq:cc} (without using training data at all). The individual bars within each subgroup average the results over 5 runs with different random seeds.

The first thing to notice is that the No-PINN solution (trained only on 120s-spaced data points) is outperformed in  every test condition.  This demonstrates the regularizing effect of the physics loss component, which improves the generalization of the model even under the same time horizon used for the training data.
The No-PINN baseline obtains a MAE of 0.068, 0.083, and 0.1 for the three different time horizons, 
being outperformed by 21\%, 22\%, and 22\% respectively by the best PINN.
Interestingly, we find that the PINN trained using all time horizons simultaneously in the physics loss (PINN-All) achieves the best performance for all test conditions, even outperforming the ones whose physics loss includes solely examples with $N_p$ equal to the tested horizon. This result suggests that giving multiple $N_p$ values to the physics equations let the network learn better the underlying relation between SoC, time, current and temperature.
In contrast, training Branch 2 using the physics loss coupled with \textit{single values} of $N_p >$ 120s does not always lead to an advantage.
Moreover, comparing the performance of PINN-All to Physics-Only reveals a consistent improvement across all scenarios. This underscores how the NN, even more when augmented with the physics component, excels at extracting valuable features from raw data, thereby significantly enhancing the model's accuracy.

\begin{figure}[t]
    \centering
    \includegraphics[width=.99\linewidth]{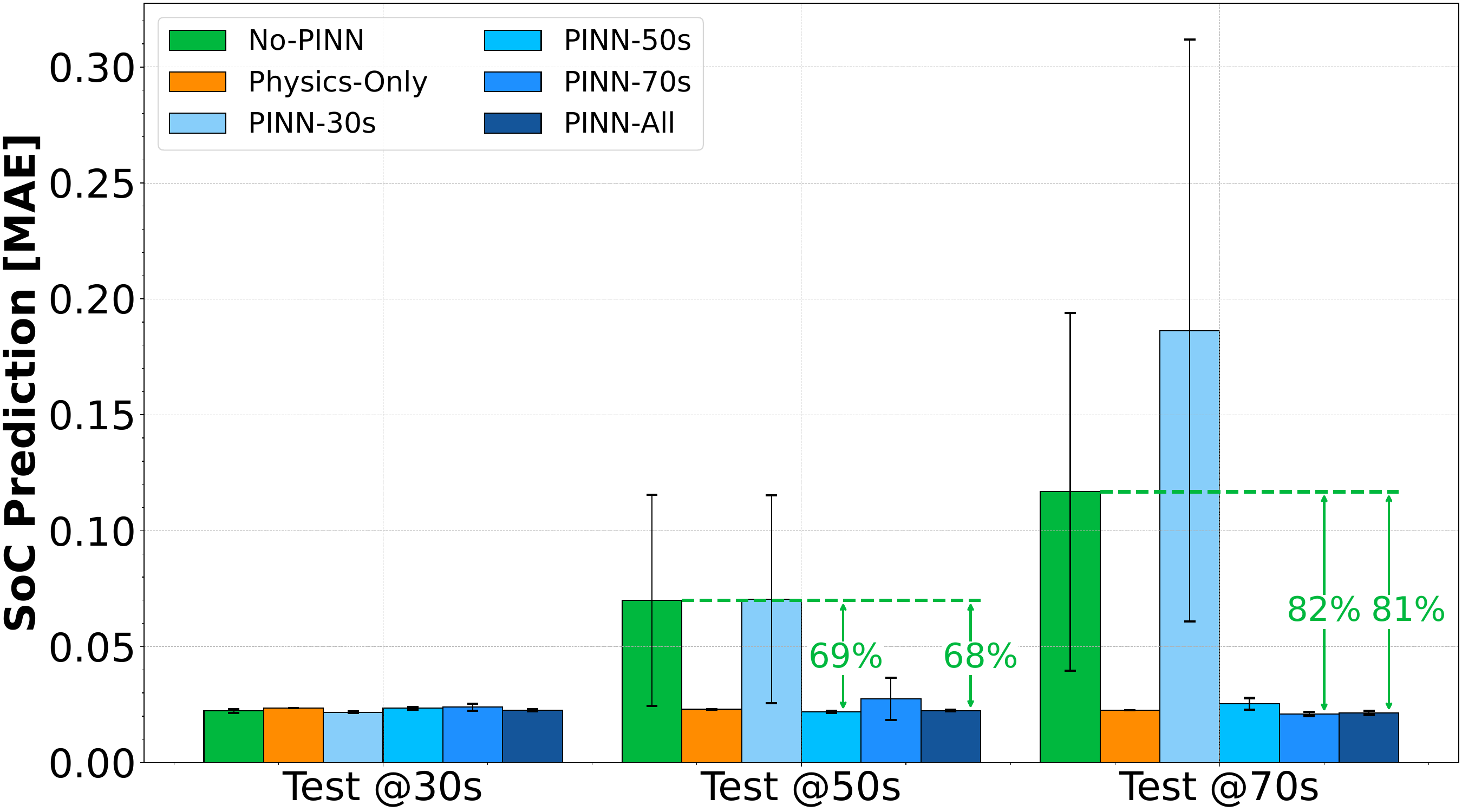}
    \vspace{-0.5cm}
    \caption{Results on LG dataset with different physic loss}
    \vspace{-0.6cm}
    \label{fig:LG}
\end{figure}
\subsection{Results on the LG Dataset}
\label{sec:lg-results}
Given the limited charge/discharge conditions (i.e., constant current) of the Sandia dataset, we also benchmark our approach on the LG dataset, which contains more complex charge and discharge patterns with varying currents.
Figure \ref{fig:LG} illustrates the result for this dataset with a visualization analogous to the previous section, but using 30, 50, and 70 seconds as prediction horizons.
It is worth observing that for this dataset, contrary to the previous one, the PINN trained with $N_p$ equal to the time horizon used for testing always achieves the best result on the corresponding test condition. 
Namely, they achieve 0.0217, 0.0218, and 0.0210 MAE on the predicted SoC, outperforming the No-PINN result by 3\%, 69\%, and 82\%, respectively.
However, similarly to the Sandia experiments, using the physics equation with multiple time horizons simultaneously (PINN-All), allows the network to closely approach the best performance in all test conditions. 
In particular, PINN-All achieves the second lowest MAE overall of 0.0214, being just 1.8\% less accurate than the best model (PINN-70s).
This shows that the physics loss allows the network to better generalize even when charge and discharge currents change over time.

\input{tables/soa}
It is also noteworthy that as the value of $N_p$ increases, the error of the No-PINN architecture grows, and the overall improvement in MAE due to physics becomes more pronounced. This confirms the importance of the physics loss in regularization, especially when dealing with data farther from the training dataset.
\vspace{-0.25cm}
\subsection{State-of-the-art Comparison}
\vspace{-0.1cm}
In this section, we focus on the LG dataset to compare our proposed approach with the best SoA algorithm for SoC estimation \cite{wong-2021} and with \cite{Dang23}, that is the most similar work to ours (as presented in Sec. \ref{sec:background}, as it also exploits physics equations during training).%

Table \ref{tab:soa} reports the results obtained on SoC estimation, i.e., the error on SoC(t), computed with data measured from the battery, and on SoC prediction, i.e., the error on SoC(t+N). 
For our method, we report the output of Branch 1 for the former, and of the full network for the latter,
using a 30s time horizon.
Note that none of the previous works tested on this dataset have considered the prediction of future SoC values. Our work addresses this more complex task, achieving an impressive SoC prediction MAE of 0.014 when exploiting physics equations to train our networks.
Moreover, we also demonstrate comparable performance on the same task of the SoA, i.e., the instantaneous SoC estimation.
When compared to \cite{wong-2021}, our approach reaches almost the same MAE (0.014 vs. 0.012) on the same test dataset, with a model that requires an impressive 260k$\times$ fewer operations and 400$\times$ less memory.
Compared to \cite{Dang23}, we obtain 4.2$\times$ lower MAE than their best solution, an LSTM with comparable dimensions of the one of~\cite{wong-2021}, and up to $\simeq$6$\times$ lower MAE when compared to their proposed MLP architecture.
We argue that we achieve better performance thanks to the introduced moving average on input windows, used as pre-processing before feeding the data to Branch 1. This allows the network to account for I, V, and T information of the last 30 seconds instead of their noisy instantaneous values.
\subsection{Full discharge analysis}
\begin{figure*}[ht]
    \centering
    \includegraphics[width=\linewidth]{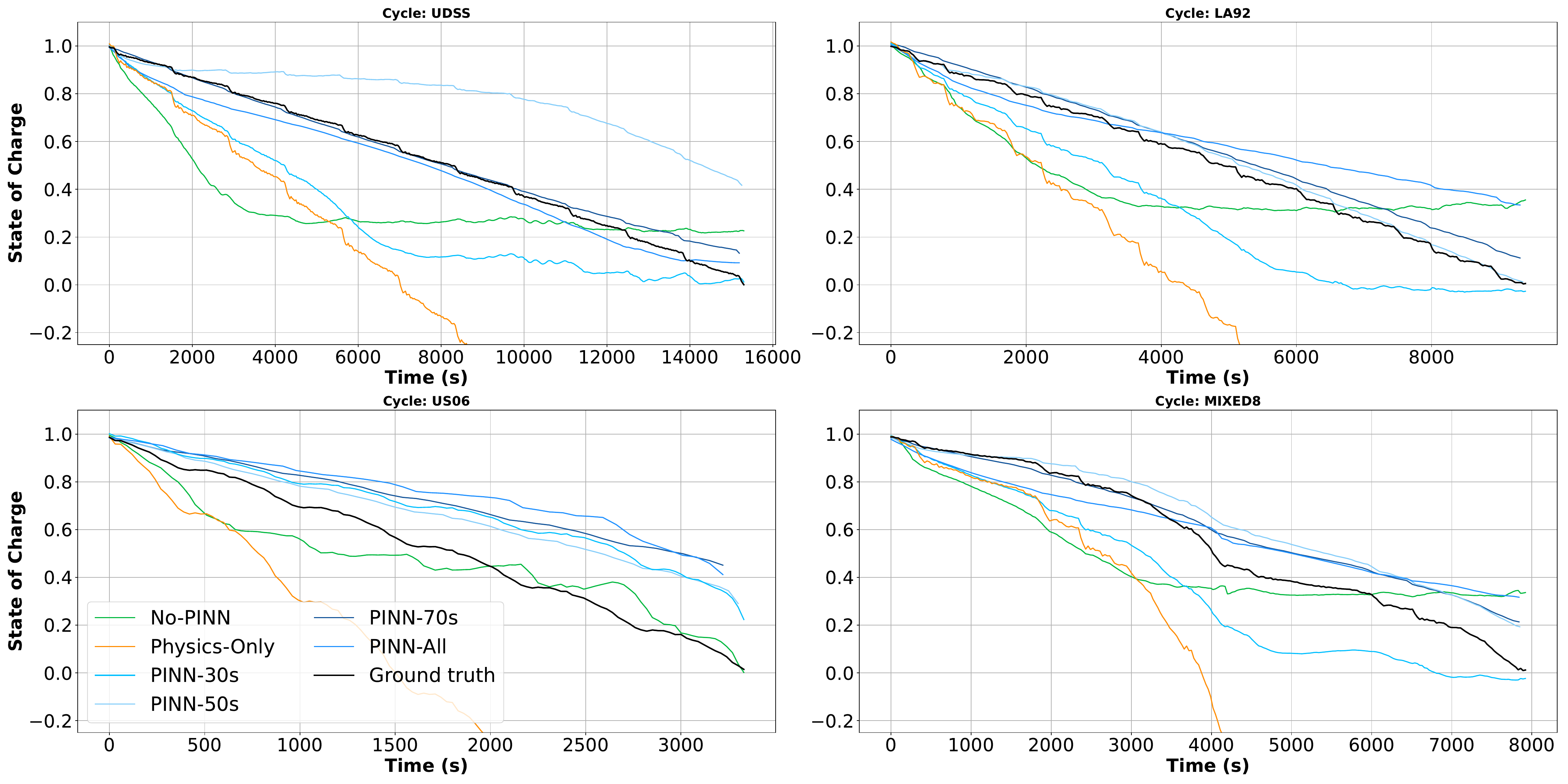}
    \caption{Auto-regressive inference with our networks, on the four driving cycles of the LG dataset, at 25°C.}
    \vspace{-0.2cm}
    \label{fig:full-discharge}
\end{figure*}
In this final section, we illustrate the estimation of a complete discharge cycle, leveraging the autoregressive application of our model, depicted in Fig.~\ref{fig:nn-autoregressive}. We target the entire four ``driving'' cycles of the LG test dataset. 
We highlight that this is a critical task, i.e., predicting the battery lifetime given a target workload, that can not be solved by SoA methods, given that they require instantaneous voltage information, whereas we use voltage as input \textit{only at the first timestamp}.
In Fig. \ref{fig:full-discharge}, each line represents a different network configuration. For all models, the single-step time horizon for SoC prediction is set as the one that resulted in the lowest MAE in the previous experiments. Then, multiple autoregressive predictions are performed. For instance, for the PINN-50s, we use 50 seconds as the time horizon: we first predict SoC(0) with Branch 1, and then recursively use Branch 2 to predict SoC(50), SoC(100), etc.
The black line represents the golden SoC of the battery. 
The No-PINN and the Physics-Only configurations use a time horizon of 30s, corresponding to the same value of $N$ present in the dataset, and to their best testing condition (see Fig. \ref{fig:LG}).
The first thing that we notice is that the No-PINN architecture achieves poor performance on SoC prediction for 3 out of 4 cycles. 
Similarly, the Physics-Only approach consistently performs the worst among all cycles. This is primarily because errors accumulate with each iteration, and without accounting for voltage in the equation, the predictions tend to diverge rapidly. Interestingly, while values are overestimated, the \textit{shape} of the discharge patterns predicted by Physics-Only closely align with the ground truth, hinting at why incorporating the physics equation during training can enhance performance.
Indeed, combining physics and data-driven training strongly improves the results, stepping from an average final SoC prediction of 0.234 (ground truth is SoC=0.0) for the No-PINN case to 0.089 for the best PINN setup, i.e., PINN-30s. 
The only cycle where the PINNs marginally underperform is the US06.
Importantly, the higher errors shown in this experiments are due to accumulation caused by the autoregressive inference,
demonstrating that this task is much more challenging compared to the estimation of the instantaneous SoC, or to the prediction of the SoC at a single future time instant.
However, we note that running an autoregressive prediction that lasts for a full discharge cycle is an extreme case, that would probably not occur in practice. A more realistic usage of our model would run a limited number of autoregressive steps, depending on the length of the workload whose effect on the SoC shall be predicted, thus, limiting the error accumulation.

%% file: tables/soa.tex
\begin{table}[]
\footnotesize
\vspace{-0.2cm}
\caption{Comparison with SoA on the LG dataset for the prediction of SoC(t) and SoC(t+N)}
\label{tab:soa}
\begin{tabular}{l|lllll}
\hline
                                 & T {[}°C{]} & SoC(t) & SoC(t+N) & Mem & Ops\\ \hline
No-PINN               & 0         & 0.031         & 0.036 & $\simeq$ 9kb &     $\simeq$ 1150    \\
No-PINN               & 25         & 0.014         & 0.016 & $\simeq$ 9kb &     $\simeq$ 1150    \\
PINN-All              & 0         & 0.031         & 0.032 & $\simeq$ 9kb  & $\simeq$ 1150       \\
PINN-All              & 25         & 0.014         & 0.014 & $\simeq$ 9kb  & $\simeq$ 1150       \\ \hline
LSTM~\cite{wong-2021}    & 25         & 0.012          & n.a.         & $\simeq$ 4Mb& $\simeq$300M  \\
LSTM~\cite{wong-2021}    & 0          & 0.017          & n.a.           & $\simeq$4Mb & $\simeq$300M\\
DE-LSTM~\cite{Dang23} & 0          & 0.129          & n.a.            & n.a. & n.a.\\
DE-MLP~\cite{Dang23} & 0          & 0.177          & n.a.            & n.a. & n.a.\\ \hline
\end{tabular}
\vspace{-0.4cm}
\end{table}

%% file: text/05-conclusions.tex
\section{Conclusions}
\label{sec:conclusions}
The \textit{prediction} of future battery SoC for different time horizons is a topic not as popular as the \textit{instantanous} SoC estimation. It requires models including as parameters the future expected workload for which one wants the SoC to be predicted.
In this work we proposed one such model, which relies on: i) a novel, two-stage NN that decouples the problem in two steps (current SoC estimation first, and then future SoC prediction using the expected workload as input); ii) the integration of a very simple yet effective equation that describes SoC evolution over time into the learning process, making the above NN \textit{physics-informed}. 
The proposed method shows accuracy comparable with the SoA (0.014 vs 0.012 MAE) with a much simpler architecture (409$\times$ fewer parameters and 260k$\times$ less operations), and, more importantly, demonstrates excellent future SoC prediction capability even for time horizons that differ from those available in the training dataset.